\documentclass[11pt]{article}

\usepackage{acl}

\usepackage{times}
\usepackage{latexsym}
\usepackage[T1]{fontenc}
\usepackage[utf8]{inputenc}
\usepackage{microtype}
\usepackage{inconsolata}
\usepackage{graphicx}
\usepackage{booktabs}
\usepackage{array}
\usepackage{tabularx}
\usepackage{url}
\usepackage{amsmath}
\usepackage{xcolor}
\usepackage{tikz}

\AtBeginDocument{%
  \setlength{\abovedisplayskip}{6pt plus 2pt}%
  \setlength{\belowdisplayskip}{6pt plus 2pt}%
  \setlength{\abovedisplayshortskip}{3pt plus 1pt}%
  \setlength{\belowdisplayshortskip}{3pt plus 1pt}%
}

\title{Co-Evolution in Agentic Systems: Toward Self-Directed Evolution  \\ Beyond Human Design}

\author{
\textbf{Qing Zong\textsuperscript{1}},
\textbf{Jiayu Liu\textsuperscript{2}},
\textbf{Junhao Shen\textsuperscript{3}},
\textbf{Zecong Tang\textsuperscript{4}}, 
\textbf{Linsi Wu\textsuperscript{5}}, 
\textbf{Yuxuan Liu\textsuperscript{1}},
\textbf{Rui Wang\textsuperscript{1}},  \\
\textbf{Zhaowei Wang\textsuperscript{1}},
\textbf{Weiqi Wang\textsuperscript{1}},
\textbf{Cheng Qian\textsuperscript{2}}, 
\textbf{Xiusi Chen\textsuperscript{3}}, 
\textbf{Yangqiu Song\textsuperscript{1}}\\ 
$^{1}$Hong Kong University of Science and Technology ~~$^{2}$University of Illinois Urbana-Champaign \\
 ~~$^{3}$The Chinese University of Hong Kong ~~$^{4}$The University of Hong Kong  ~~$^{5}$Peking University\\
\texttt{\{qzong, yqsong\}@cse.ust.hk}
}

\begin{document}
\maketitle

\begin{abstract}
Agentic systems are increasingly expected to improve after deployment, yet single-entity self-evolution is often bounded by a static learning context, such as fixed tasks and feedback. This survey focuses on co-evolution in agentic systems, a multi-component form of self-evolution in which multiple agents and their environment impose adaptive pressure on one another. To organize existing papers, we propose a progressive three-stage taxonomy that traces how the system gradually sheds human-engineered constraints. Agent--Agent Co-Evolution studies how agents adapt through dynamic peers, including adversarial, collaborative, and organizational adaptation. Agent--Environment Co-Evolution extends this loop to adaptive tasks, feedback, and interaction spaces that change with the agents. Meta Co-Evolution further explores the possibility of making the evolution mechanism itself evolvable. We also discuss open challenges in evaluating such systems, scaling them across multiple components, and keeping increasingly autonomous evolutionary processes safe and controllable. This survey provides a unified foundation for building robust and open-ended agentic systems that can improve beyond fixed human-designed paths.
\end{abstract}

\section{Introduction}

Recent advances in AI have shifted attention from isolated models to agentic systems, which can autonomously interact with other agents and external environments~\citep{guo2024llmmultiagents,plaat2025agenticllm,luo2025llmagent}. They may be augmented with a harness~\citep{ning2026codeharness,yao2025harness}, including tools~\citep{xu2025toollearningagents}, memory~\citep{du2026memoryagents}, and skills~\citep{zhou2026agentskills}. As agentic systems become increasingly complex, a central question is how they can continue to improve after deployment~\citep{gao2025selfevolvingagents}.

\begin{figure}[t]
    \centering
    \includegraphics[width=\columnwidth]{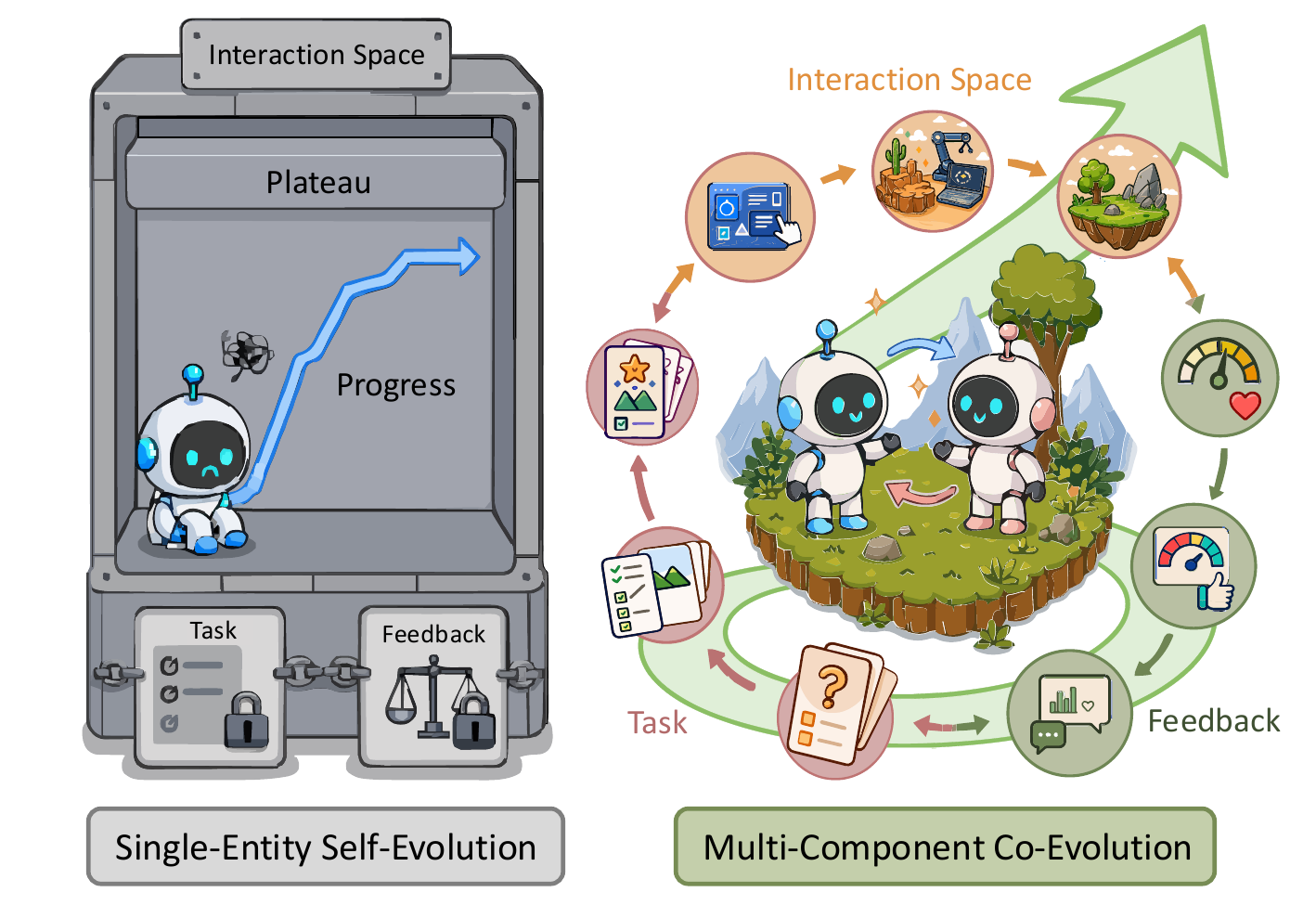}
\caption{Comparison between single-entity self-evolution and multi-component co-evolution.}
    \label{fig:fig1}
\vspace{-0.1in}
\end{figure}

Self-evolution is a major paradigm for continual improvement in agentic systems, where an agent persistently updates itself from experience, feedback, and failures without human intervention. One important form is single-entity self-evolution, where updates occur within one agent, such as refining its model backbone~\citep{zhai2025agentevolver,wu2025evolver}, updating memory~\citep{suzgun2025dynamiccheatsheet}, or improving skills~\citep{xiao2026socraticswe}.
However, it can remain bounded by fixed external conditions, as illustrated in Figure~\ref{fig:fig1}. This limitation echoes the Red Queen effect~\citep{vanvalen1973redqueen}: sustained progress requires mutual adaptation rather than adaptation on only one side. 

\begin{figure*}[t]
\centering
\includegraphics[width=\textwidth]{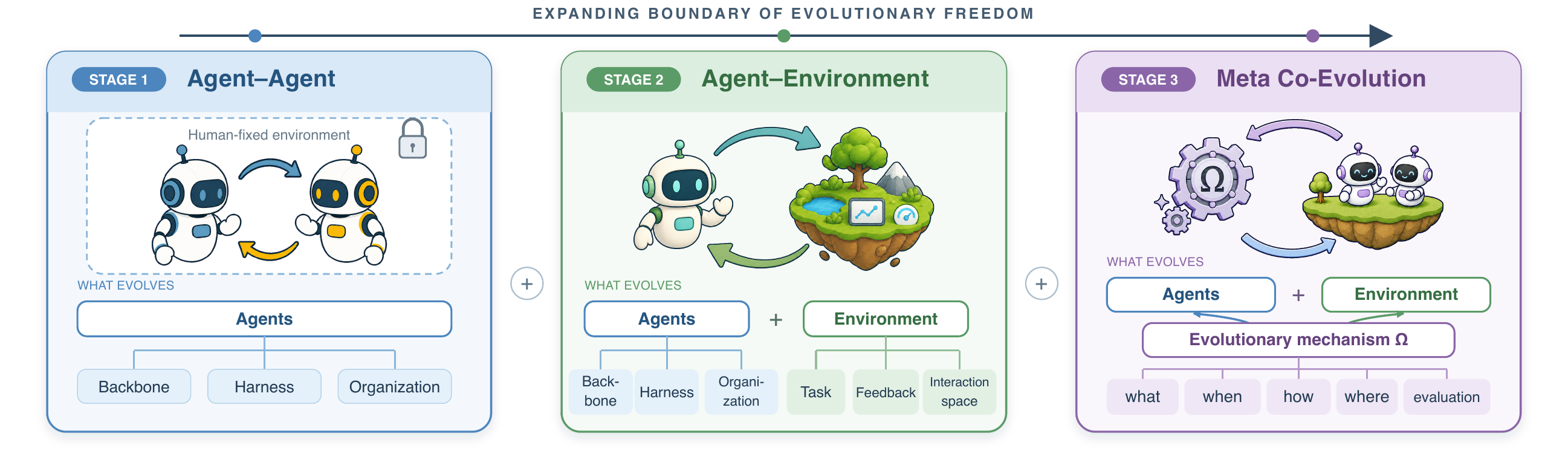}
\caption{The progressive taxonomy of co-evolution, which reflects an expanding boundary of evolutionary freedom.}
\label{fig:stages}
\end{figure*}

Co-evolution extends self-evolution by making the counterpart adaptive~\citep{hillis1990coevolving}. In agentic systems, agents may evolve with adaptive peers such as opponents or collaborators~\citep{pan2026coverrl}, or with environments such as tasks, feedback, and interaction spaces~\citep{guo2025genenv,liu2026comap}. Recent work further suggests a meta-level extension in which the evolution mechanism itself becomes evolvable~\citep{iacob2026redqueen}. Across these settings, multiple components jointly adapt and continually reshape each other's further evolution, rather than merely exchanging information or interacting.

Despite growing interest, no survey has centered on such co-evolutionary agentic systems. They examine LLM-based agents and multi-agent systems~\citep{guo2024llmmultiagents,plaat2025agenticllm,luo2025llmagent}, agent components and harness engineering~\citep{xu2025toollearningagents,du2026memoryagents,zhou2026agentskills,meng2026agentharness}, and environment scaling for interactive agents~\citep{huang2025envscaling}. Surveys on self-evolving agents~\citep{gao2025selfevolvingagents,xiang2026systematicselfevolving} are closest to our scope, but treat co-evolution as a subtheme rather than the central organizing axis. We address this gap by asking which components co-evolve, how the boundary of evolutionary freedom expands across components, and what the ultimate form of this process might be.

To organize this literature, we propose a progressive three-stage taxonomy structured by the expanding scope of what a system is allowed to evolve, tracing how a system gradually sheds human-engineered constraints. (1) \textbf{Agent--Agent Co-Evolution} captures mutual adaptation among evolving peers within a fixed environment, including changes to their model backbones, harness components, and organizational structures. (2) \textbf{Agent--Environment Co-Evolution} extends adaptation to environmental components, including tasks, feedback, and interaction spaces. (3) \textbf{Meta Co-Evolution} captures an emerging stage in which the evolution mechanism itself becomes adaptive, allowing the system to revise what, when, how, and where to evolve, as well as how to evaluate. This progression expands evolutionary freedom from agents alone, to agents and environments, and ultimately to the rules governing their evolution, providing a possible pathway toward open-endedness~\citep{stanley2017open,hughes2024open}, where systems keep improving rather than converging to a fixed endpoint.

This survey makes three contributions. (1) to our knowledge, we offer the first focused survey of co-evolution in agentic systems, distinguishing mutual adaptation from mere interaction or information exchange. (2) we organize the literature with a progressive three-stage taxonomy that traces how evolutionary freedom expands from agents alone, to agents and environments, and ultimately to an evolvable evolution mechanism. (3) we discuss open challenges and future directions for evaluating, scaling, and governing increasingly autonomous co-evolutionary systems.

The remainder of this paper is organized as follows.
\S~\ref{sec:foundations} defines formal foundations and introduces our three-stage progressive taxonomy.
\S~\ref{sec:stage1}--\ref{sec:stage3} review methods across the three stages.
\S~\ref{sec:discussion} discusses challenges and future directions.
\S~\ref{sec:conclusion} concludes.

\section{Preliminaries}
\label{sec:foundations}

\subsection{What Is an Agentic System?}
\label{sec:foundations-agent}

An agentic system $S=(A,\,E)$ consists of an agent collective $A$ and an environment $E$. An agent is a system unit that can autonomously interact with other agents or the environment. To track evolution, we treat each agent as a unified evolving unit and denote it by $a_i=(m_i,\,h_i)$: a model backbone $m_i$ and a harness $h_i$~\citep{yao2025harness}, such as memory, tools, skills, prompts, or workflows. In practice, $h_i$ may include only some of these components, or none. Agents may share the same backbone yet remain distinct if their harnesses instantiate different roles or objectives. An agent evolves when either component changes:
\[
    \Delta a_i \neq 0 \iff (\Delta m_i \neq 0) \lor (\Delta h_i \neq 0).
\]

Agents are not an unordered set. They are structured by $\Pi$, which encodes their roles, communication topology, and division of labor:
\[
    A = \big(\{a_1, \dots, a_n\},\; \Pi\big).
\]
$A$ evolves whenever an $a_i$ or $\Pi$ changes.

The environment $E$ is everything external to $A$ that agents act upon and receive feedback from. A trajectory $\tau^t$ records the sequence of thought--action--observation cycles~\citep{yao2023react} accumulated by the system up to step $t$.

\subsection{What Is Co-Evolution?}
\label{sec:foundations-coevo}

Let $\Omega$ be the evolution mechanism driving state transitions, written as $S^{t+1}=\Omega(S^t,\tau^t)$. It specifies what can evolve, when evolution is triggered, how variants are generated or updated, where evolution takes place, and how evolution quality is evaluated. In this survey, co-evolution is a coupled form of self-evolution in which at least two evolving units jointly adapt and continually reshape each other's further evolution, rather than merely exchanging information or interacting. It requires:
\[
\begin{aligned}
S^{t+1}&=\Omega(S^t,\tau^t),\quad \exists\,x\neq y\in S^t:\\
x^{t+1}&\neq x^t,\quad y^{t+1}\neq y^t,\\
x&\overset{\text{evolutionary pressure}}{\longleftrightarrow}y.
\end{aligned}
\]

\subsection{Three-Stage Taxonomy}
\label{sec:foundations-taxonomy}

Our three-stage taxonomy (Figure~\ref{fig:stages}) follows the expanding scope of what a system is allowed to evolve. From one stage to the next, human-engineered constraints are gradually removed.

\paragraph{Stage~1 --- Agent--Agent Co-Evolution.}
Co-evolution begins when an agent no longer learns against a static counterpart. As agents respond to one another, each agent's progress changes the challenges and opportunities facing its peers, and the collective may also reorganize its structure $\Pi$. This creates a coupled evolutionary process within the agent collective:
\[
    A^{t+1} = \big(\{a_i^{t+1}\}_{i=1}^{n},\; \Pi^{t+1}\big)
        = \Omega(A^t,\; E,\; \tau^t).
\]

\paragraph{Stage~2 --- Agent--Environment Co-Evolution.}
This coupled process can continuously change the agents, but not the environment in which they interact. When the environment remains fixed, it limits the new experiences and pressures that the agents can encounter. Stage~2 therefore extends the process to the environment itself, allowing agent behavior to reshape the whole conditions that subsequently shape the agents:
\[
    (A^{t+1}, E^{t+1}) = \Omega(A^t,\; E^t,\; \tau^t).
\]

\begin{figure*}[t]
  \centering
  \includegraphics[width=\textwidth]{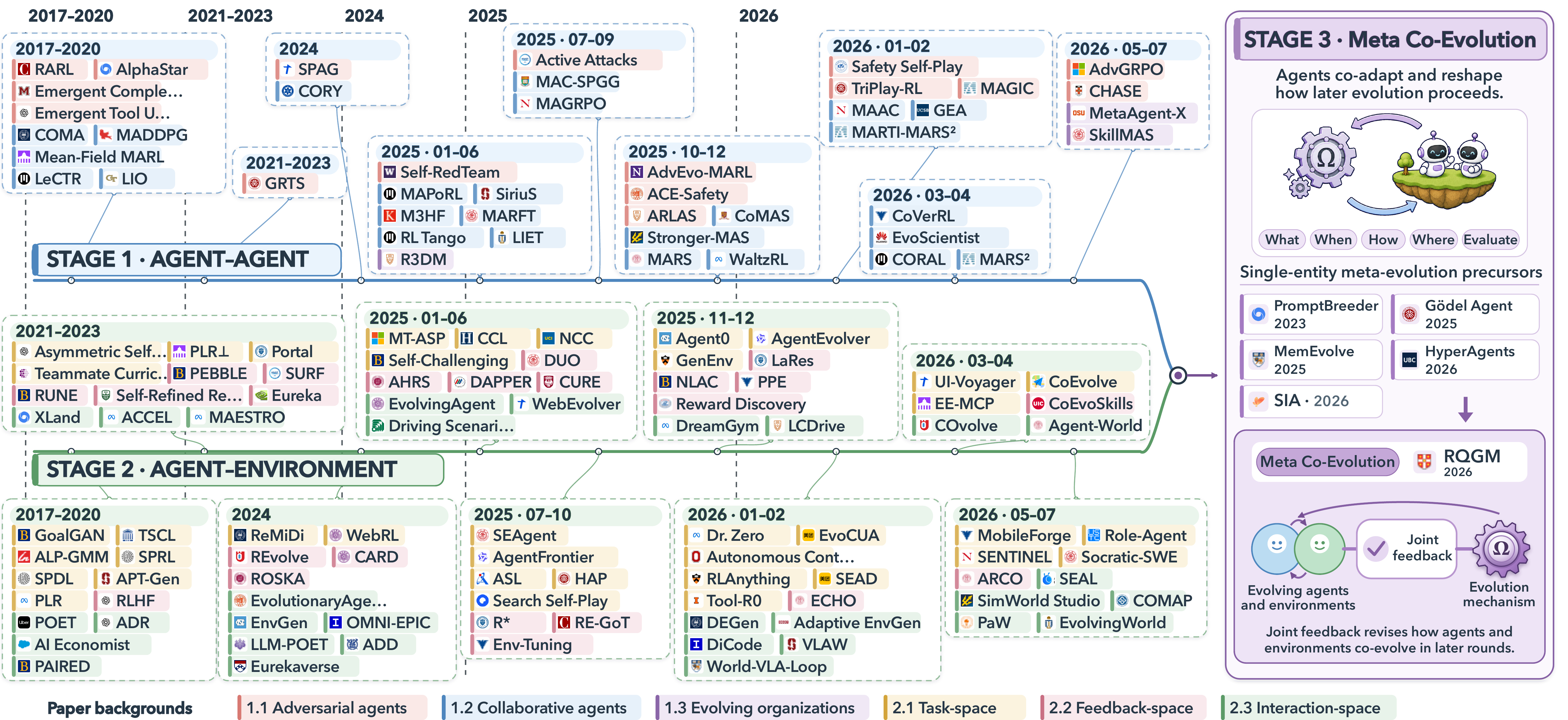}
  \caption{Paper landscape of co-evolution in agentic systems.}
  \label{fig:paper-landscape}
\end{figure*}

\paragraph{Stage~3 --- Meta Co-Evolution.}
In Stage~2, the evolution mechanism remains human-designed, so the system can change its contents but not how they change. We therefore define \emph{meta co-evolution} as a stage where the lower-level co-evolving system further revises its evolution mechanism through a self-generated revision process $\Gamma^t$:
\begin{align*}
    \Omega^{t+1} &= \Gamma^t(S^t,\; \Omega^t,\; \tau^t), \\
    S^{t+1} &= \Omega^{t+1}(S^t,\; \tau^t).
\end{align*}
This recursion provides a pathway toward open-endedness~\citep{stanley2017open,hughes2024open}, where the system continues to generate meaningful novelty rather than converging to a fixed endpoint, characterized by continuous novelty, $\Omega^{t+1}\neq\Omega^t$, and unbounded divergence, with adaptive capability $\mathcal{H}$ satisfying $\lim_{t\to\infty}\mathcal{H}(S^t,\Omega^t)=\infty$.

\section{Agent--Agent Co-Evolution}
\label{sec:stage1}

This stage studies co-evolution within the agent collective $A=(\{a_i\}_{i=1}^{n},\Pi)$, where each agent's main source of improvement is the other evolving agents around it. We organize this stage into three patterns: adversarial and collaborative co-evolution distinguished by agents' goal relations, and evolving agent organizations that further adapt agents' roles and interaction structures.

\subsection{Adversarial Agents}
\label{sec:stage1-adversarial}
Adversarial agents co-evolve through opposing objectives, where each side is rewarded for defeating the other, so one agent's progress directly raises the difficulty its opponent faces. We organize this literature by how the pressure scales, from pairwise pressure between two evolving sides to multi-source pressure across many adversarial agents.

\subsubsection{Pairwise Adversarial Pressure}
This co-evolution arises between two adversarial agents with distinct roles, and can trace back to the GAN~\citep{goodfellow2014gan}, where a generator produces fake images and a discriminator learns to tell them from real ones, each improving with the other. The same idea was soon used for embodied control. RARL~\citep{pinto2017robust} trains a control policy against an adversary that adds destabilizing forces, making it robust to real disturbances rather than a fixed noise pattern. Through competitive self-play in fighting games~\citep{bansal2018emergent} and hide-and-seek~\citep{baker2019emergent}, agents learn tool use and counter-strategies as they escalate against each other. The idea later moved to language, where SPAG~\citep{cheng2024selfplaying} trains both agents of a word-guessing game, with one trying to make the other say a secret word while the other tries to guess it.

Most recently, this idea has been widely applied to LLM safety, where an attacker agent generates jailbreak prompts and a defender agent learns to refuse them, each driving the other to improve. ACE-Safety~\citep{ace2025safety} searches for attacks with Monte Carlo tree search, AdvGRPO~\citep{advgrpo2026learning} stabilizes joint GRPO with dense multi-channel rewards, and \citet{activeattacks2025} makes already-found attacks stop working, so the attacker keeps discovering new attack types instead of collapsing to a few modes. MAGIC~\citep{magic2026coevolving} runs a multi-turn game, so the attacker can find weaknesses that single-turn attacks cannot, and CHASE~\citep{chase2026adversarial} drops preset attack templates so its defender stays robust to unseen attacks. Some works also merge both roles into one model through self-play, where Self-RedTeam~\citep{selfredteam2025moving} keeps both online against each other and \citet{beyourown2026redteamer} replays past failures to avoid forgetting hard cases. Beyond direct prompts, ARLAS~\citep{arlas2025adversarial} extends the game to tool use, hiding harmful instructions in the tool outputs an agent reads. 

\subsubsection{Multi-Source Adversarial Pressure}
Multi-source adversarial pressure arises when an agent must remain competitive against multiple opponents. AlphaStar~\citep{vinyals2019alphastar} provides a classical example through league training in a real-time strategy game, main agents co-evolve with exploiters that target their weaknesses and past versions that prevent forgetting.

Later works extend this idea in different ways. \citet{redteamgame2024} co-evolve a defender against a diverse population of red-team models, so it generalizes across attack styles instead of overfitting to a single attacker. TriPlay-RL~\citep{triplay2026} co-evolves three agents, an attacker, a defender, and an evaluator, so that attack generation, safe response, and safety judgment evolve together. AdvEvo-MARL~\citep{advevo2025shaping} trains attackers against a whole team of task agents, which learn to resist attacks while doing their jobs, so safety is built into the group rather than a separate guard.

\subsection{Collaborative Agents}
\label{sec:stage1-collaborative}
Unlike adversarial agents, collaborative agents co-evolve through shared goals, where one agent's improvement forces its partners to adjust what they coordinate with and build on. We distinguish two structures, parallel collaboration, where equivalent agents either have no roles or can take any role, and role-differentiated collaboration, where fixed and distinct roles give agents different paths of improvement.

\subsubsection{Parallel Collaboration}
Parallel collaboration originates in Multi-Agent Reinforcement Learning (MARL), where peer agents share a single task reward and co-adapt as the others chang~\citep{lowe2017maddpg,foerster2018coma,yang2018meanfieldmarl}. Recent work brings this to LLM agents. \citet{llmcollabmarl2026} trains them with shared rewards and group-relative advantages, while \citet{decentralizedllmactorcritic2026} compares joint-history and local-history critics to stabilize their co-evolution. MARS$^2$~\citep{mars22026,martimars22026} lets multiple agents co-evolve on a shared search tree, spreading each program's test reward across the tree instead of only its final node.

Beyond the final outcome, other work draws co-evolutionary pressure from the collaboration process itself~\citep{lectr2018,incentivize2020}. Some methods turn interaction traces into rewards, where CoMAS~\citep{comas2025} and MAPoRL~\citep{maporl2025} reward agents from their discussion, such as whether a turn corrects or persuades later responses. Others revise the signals that guide later cooperation, where LIET~\citep{liet2025} builds a shared cooperation-tip list from messages, M3HF~\citep{m3hf2025} converts feedback on team behavior into per-agent rewards, and MAC-SPGG~\citep{macspgg2025} uses a sequential public-goods reward to discourage free-riding.

Parallel collaboration can also span separate workspaces, where agents explore independently and share what they learn, as in CORAL~\citep{coral2026}, which diffuses attempts, notes, and skills through shared memory, and GEA~\citep{gea2026}, which treats the group as the evolutionary unit so that execution logs, failures, and improvement from one agent guide the updates of others.

\subsubsection{Role-Differentiated Collaboration}
Role-differentiated collaboration co-evolves agents with distinct roles. A common pattern is a produce-and-revise loop. CORY~\citep{cory2024} is a minimal form where a pioneer answers and an observer revises, with periodic role exchange keeping the two coupled. RL Tango~\citep{zha2025rltango} trains the verifier from outcome correctness to give process feedback, while CoVerRL~\citep{pan2026coverrl} bootstraps its verifier from majority-vote answers when labels are unavailable. WaltzRL~\citep{waltzrl2025} jointly trains a conversation agent and a critique agent, rewarding it only when its suggestions improve the partner's next response. Others split tasks by agent capability. EvoScientist~\citep{lyu2026evoscientist} lets researcher and engineer agents accumulate experience across tasks through persistent memories. SiriuS~\citep{sirius2025} trains domain agents such as physicist, mathematician, and summarizer from successful and repaired collaboration trajectories, while MARS~\citep{marsdeep2025} couples System1 information compression with System2 reasoning through shared trajectory rewards. Finally, MARFT~\citep{marft2025} and Stronger-MAS~\citep{strongermas2026} focus on credit assignment, attributing the shared team reward to each role so every role can improve.

\subsection{Evolving Agent Organizations}
\label{sec:stage1-organizations}
Beyond fixed roles, agents can co-evolve with the organization, including role assignments and interaction structure. R3DM~\citep{r3dm2025} discovers roles from agent behavior and co-adapts them with agent policies, pushing agents toward distinct behaviors. SkillMAS~\citep{skillmas2026} jointly updates agent skills and team structure, restructuring it when execution traces reveal a mismatch. MetaAgent-X~\citep{zhang2026metaagentx} trains the workflow designer and executor agents together so they improve each other.

\section{Agent--Environment Co-Evolution}
\label{sec:stage2}
This stage studies co-evolution between agents and an adaptive environment $E$. As agents improve, the environment reshapes the tasks they face, the feedback on their outputs, or the worlds they act in, and we classify methods by which change primarily drives further agent evolution.

\subsection{Task-Space Co-Evolution}
\label{sec:stage2-task-space}
In task-space co-evolution, the environment adapts what problem the agent solves next. This literature takes two forms, selecting tasks from an existing pool and generating new task specifications.

\subsubsection{Exposure and Selection}
Exposure and selection methods keep the task space fixed and adapt which tasks the agent sees. Early work adjusted exposure by learning progress. For discrete tasks or data sources, curriculum learning~\citep{matiisen2017teacher} adjusts sampling weights by estimated progress, while methods over continuous task parameters~\citep{klink2020selfpacedcontextual,klink2020selfpaceddeep,portelas2020teacher,florensa2021aptgen} steer sampling toward the agent's current competence, an idea \citet{heterogeneous2025} later extends with a teacher model that raises the sampling of tasks the student still fails. Beyond competence matching, replay-based selection~\citep{jiang2021prioritized,jiang2021replayguided} revisits past levels that still offer learning potential, and regret-based selection~\citep{parkerholder2024remidi,uedoptimization2025} prioritizes configurations where the agent lags behind a stronger reference. In multi-agent settings, \citet{clcooperation2023} control task difficulty through changing the exposure to teammates with different capabilities. PORTAL~\citep{portal2024} selects intermediate tasks that lead agents to a hard target step by step. For LLM agents, \citet{roleagent2026} resamples tasks matching the agent's own failure modes, and SEAD~\citep{sead2026} samples simulated user profiles that become harder for the service agent to handle as it improves.


\subsubsection{Adaptive Task Generation}
Generative task construction creates new tasks rather than reshuffling an existing pool, placing them near the agent's current competence frontier~\citep{goalgan2018}. Even when a task is produced by an agent, it belongs to environment evolution, because that agent is not a peer competing or collaborating within the same task.

Embodied work makes this explicit through goal generation. \citet{sadykov2021asymmetric} explores environment for challenging yet reachable goals, and \citet{doasyouteach2025} broaden goal coverage with multiple teachers. CCL~\citep{ccl2025} extends this to multi-agent RL with agent-specific subtasks.

Recent LLM-agent methods instantiate this loop across several interactive settings. In tool use, generators calibrate task difficulty to the solver's success rate~\citep{guo2025genenv,acikgoz2026toolr0,agent02025,agentfrontier2025}, convert its failed rollouts into new tasks~\citep{sentinel2026,coevolve2026,agentevolver2025}, or pair each task with a verification function~\citep{selfchallenging2025}.
In search tasks, proposers generate multi-hop verifiable questions from retrieval behavior~\citep{searchselfplay2025,asl2025,yue2026drzero}. In GUI and computer-use, some methods explore unfamiliar software to learn what operations it supports and turn these into tasks~\citep{liu2026mobileforge,xue2026acurl,sun2025seagent}, others derive tasks from the agent's failed attempts~\citep{webrl2024} or by perturbing parameters such as time and quantity in existing tasks~\citep{lin2026uivoyager}, and others~\citep{xue2026evocua,he2026eemcp} generate the validators and setup scripts that each task needs to run. In software engineering, Socratic-SWE~\citep{xiao2026socraticswe} builds repository repair tasks from its solving traces. Finally, RLAnything~\citep{rlanything2026} jointly rewrites tasks, policy, and reward across computer-use, game, and coding tasks.

\subsection{Feedback-Space Co-Evolution}
\label{sec:stage2-feedback-space}
Beyond adapting what tasks agents face, the environment can also co-evolve in how it evaluates, rewards, or diagnoses agent behavior.

\subsubsection{Preference-Driven Feedback}
Feedback co-evolution may be driven by an evolving reward model that learns from preference comparisons~\citep{christiano2017preferences} over trajectory pairs, scored by a scripted teacher rather than a human. PEBBLE~\citep{pebble2021} relabels replay-buffer trajectories whenever the reward model changes, reusing stored rollouts instead of collecting fresh labels. To further cut labeling, \citet{rewarduncertainty2022} and \citet{ppe2025} seek pairs where the reward model is least confident, DAPPER~\citep{dapper2025} pairs trajectories from different policies to widen behavioral contrast, and DUO~\citep{duo2025} targets pairs with strong reward disagreement and diverse segment behavior.

\begin{figure*}[t]
  \centering
  \includegraphics[width=\textwidth]{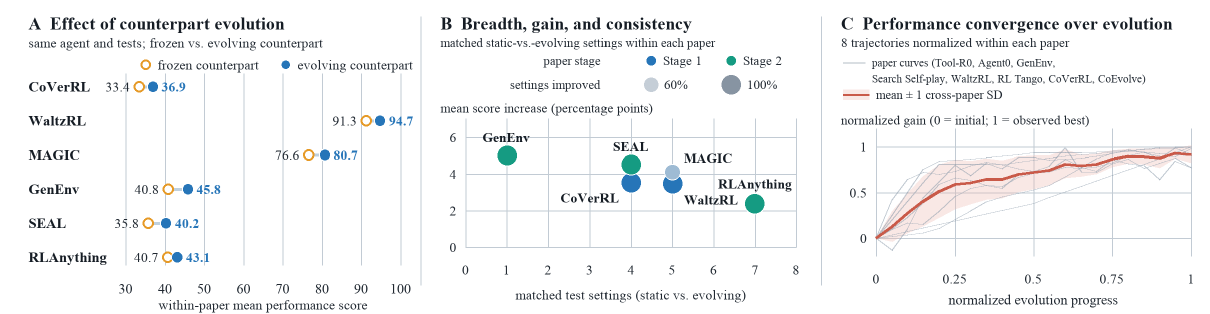}
  \caption{Cross-paper evidence for the effect, consistency, and convergence of co-evolution. See Appendix~\ref{app:cross-paper-evidence} for data selection and aggregation details.}
  \label{fig:cross-paper-evidence}
\end{figure*}

\subsubsection{Outcome-Driven Feedback}
Feedback co-evolution may also be driven by task outcomes. In embodied control~\citep{selfrefinedreward2023,eureka2023,card2024,revolve2024}, LLM-based systems train a policy under a candidate reward function and revise that function from trajectory returns or task failures. ROSKA~\citep{rewardpolicyskill2025} and LaRes~\citep{lares2025} co-search reward candidates and policy variants, keeping the best-performing rewards, RE-GoT~\citep{regot2025} refines decomposed rewards from rollout feedback including visual evaluation, and \citet{rewarddiscovery2025} updates reward parameters from policy regret. CURE~\citep{wang2025cure} and CoEvoSkills~\citep{coevoskills2026} evolve their unit tests or surrogate verifiers from execution failures. Beyond rewriting rewards or verifiers, \citet{envtuning2026} adds corrective hints early in training, then withdraws them as the agent improves, while AHRS~\citep{ahrs2025} reweights fixed reward terms by their current contribution to learning.

\subsubsection{Consistency-Augmented Feedback}
Consistency constraints further augment evaluator learning on top of preference or outcome signals. SURF~\citep{surf2022} augments sparse labels by reusing segment pairs the reward model already ranks confidently. R*~\citep{rstar2025} updates reward parameters only when several critics agree on the same segment ranking. ARCO~\citep{arco2026} co-trains a step-scoring rubric evaluator with the policy and requires those step scores to add up to the final task outcome. NLAC~\citep{nlac2025} trains a language critic so that its predictions of what comes next stay consistent with later steps on the trajectory, and ECHO~\citep{echo2026critics} updates its diagnostic critic only when acting on its advice actually improves the policy.

\subsection{Interaction-Space Co-Evolution}
\label{sec:stage2-interaction-space}
Beyond tasks and feedback, the environment can also adapt where the agent acts in, either constructing an increasingly challenging executable world or learning a world model to replace the real one.

\subsubsection{Executable World Construction}
In classical gridworld navigation and continuous control, POET~\citep{poet2019} evolves a population of environment-agent pairs, while regret-based methods~\citep{paired2020,accel2022,add2024} generate levels that target the agent's weaknesses. More recently, DEGen~\citep{degen2026} generates level parts online as the student explores, and COvolve~\citep{covolve2026} uses an LLM to co-evolve the code for both levels and agent controllers.

In robotics and embodied simulation, ADR~\citep{adr2019} widens simulation randomization as the policy improves, while later work generates worlds directly. OMNI-EPIC~\citep{omniepic2024} filters new environments by interestingness, and LLM-POET~\citep{llmpoet2024} adapts POET to embodied simulation with an LLM generator, Others~\citep{eurekaverse2024,drivingscenarios2025,adaptiveenvgen2026} generate environments whose difficulty tracks the agent's learning potential, and SimWorld Studio~\citep{simworldstudio2026} builds 3D scenes validated by physics and vision checks.

In games and open-ended worlds, XLand~\citep{xland2021} adapts game rules and layouts to agent performance. MAESTRO~\citep{maestro2023} extends regret-based design to multi-agent games by generating levels around different co-players. EnvGen~\citep{envgen2024} regenerates environments targeting the agent's weak skills, and DiCode~\citep{dicode2026} grows a level archive, generating new levels from the highest-value ones.

For tool-use agents, Agent-World~\citep{agentworld2026} expands tool combinations and database states to expose new capability gaps, while SEAL~\citep{hu2026seal} refines the environment's constraints and recovery feedback from verified rollouts to correct recurring agent errors.

Finally, construction targets the rules governing agent populations. \citet{aieconomist2022} learns a tax policy that reshapes the incentives of economic agents, and \citet{agentnorms2024} evolve social norms that reshape agent behavior across generations.

\subsubsection{Model-Based World Construction}

Recent work replaces real environments with world models that co-evolve with agents. In text-based settings, WebEvolver~\citep{webevolver2025} co-trains a web world model with the agent policy as a virtual web server for synthetic rollouts and look-ahead. COMAP~\citep{liu2026comap} uses an on-policy textual world model to predict future feedback that guides action refinement. DreamGym~\citep{dreamgym2025} synthesizes transitions from an experience model whose replay buffer grows during training. PaW~\citep{paw2026} adds next observation prediction as an auxiliary signal in agent RL. In role-play simulation, EvolvingWorld~\citep{EvolvingWorld} lets the world model decide which aspects to track and evolves the world state that later shapes character evolution.
In embodied and visual domains, world models reduce costly real-world interaction. EvolvingAgent~\citep{evoagent2025} improves a world model and agent planner from Minecraft experience. \citet{worldvlaloop2026} and \citet{vlaw2026} use a video world model predicting future frames to train the agent policy, and the two improve together. LCDrive~\citep{lcdrive2026} optimizes latent world-model tokens together with driving actions through closed-loop RL.

\section{Meta Co-Evolution}
\label{sec:stage3}

\subsection{Toward Open-Endedness}

As shown in Figure~\ref{fig:cross-paper-evidence}, Stage~1 and Stage~2 improve performance across most settings, but the gains become smaller as evolution approaches a plateau. By allowing the evolution mechanism to change, meta co-evolution may move beyond this bottleneck and open new directions for improvement.

Meta co-evolution offers a route toward open-endedness~\citep{stanley2017open,hughes2024open}, which requires continuous novelty ($\Omega^{t+1}\neq\Omega^t$), so mechanism updates can keep opening new learnable directions, and unbounded divergence ($\lim_{t \to \infty} \mathcal{H}(S^t,\Omega^t)=\infty$), so the system's adaptive capacity has no finite upper bound. In this sense, meta co-evolution is not merely another category, but a step from adaptation under designed rules toward self-expanding agentic systems.

We decompose the evolution mechanism into five adaptive decisions: (1) what to evolve selects the adaptation target, such as an agent backbone, harness, or organization, or environment artifacts like tasks, rewards, and worlds; (2) when to evolve triggers updates after failures, plateaus, or distribution shifts; (3) how to evolve produces variants through backbone training, harness revision, or structural generation and editing; (4) where to evolve covers both the domain of evolution, such as tool use, web, or robotics, and the setting it runs in, such as sandboxes, simulated or real environments; and (5) how to evaluate judges evolution quality by performance, novelty, safety, or robustness, shaping the direction of later evolution. This criterion separates meta co-evolution from isolated meta-evolution. The mechanism revision must be induced by the joint trajectory of a lower-level co-evolving ecosystem and must then alter that ecosystem's subsequent evolutionary conditions.

\subsection{From Precursors to Meta Co-Evolution}
Most related methods remain single-entity precursors. PromptBreeder~\citep{promptbreeder2023} evolves mutation prompts that control how future task prompts change. G\"odel Agent~\citep{godelagent2025} and HyperAgents~\citep{hyperagents2026} evolve the agent's own self-modification machinery. MemEvolve~\citep{memevolve2025} evolves the memory architecture that governs experience reuse, while SIA~\citep{hebbar2026sia} uses trajectory feedback to choose between harness and weight updates. These methods make the evolution mechanism evolvable but lack a lower-level co-evolving system.

RQGM~\citep{iacob2026redqueen} goes beyond by co-evolving task agents and evaluators, while its meta-agent uses joint feedback to guide later evolution.

\section{Challenges and Future Directions}
\label{sec:discussion}

\paragraph{Dynamic evaluation.}
Existing benchmarks mainly measure an agent's final capabilities across domains such as tool use~\citep{yao2024taubench,lu2025toolsandbox,liu2026planbench,liu-etal-2026-costbench,liu2026adaplanbench}, web browsing~\citep{zhou2023webarena,wei2025browsecomp}, software engineering~\citep{jimenez2024swebench,miserendino2025swelancer}, computer use~\citep{trivedi2024appworld,xie2024osworld,terminalbench2026}, and multi-agent interaction~\citep{zhou2024sotopia,mou2025agentsense,zhu2025multiagentbench}. PostTrainBench~\citep{rank2026posttrainbench} takes a step toward evolution-aware evaluation by testing whether an agent can autonomously post-train a base LLM. However, evaluating co-evolution requires going further: it must determine whether all evolving components improve, whether their gains transfer to unseen partners and environments, and how each component contributes to the joint progress. Since higher task success can hide exploitative behavior~\citep{thaman2026reward}, such systems may fail through evaluator exploitation, partner overfitting, or diversity collapse. Evaluation should therefore pair fixed benchmarks with process-level testing, such as historical cross-play, component ablations, and held-out evaluators.

\paragraph{Scaling co-evolution.}
Current systems mostly study local loops, such as attacker--defender training~\citep{magic2026coevolving}, policy--reward adaptation~\citep{rewarddiscovery2025}, or agent--task generation~\citep{acikgoz2026toolr0}. Future work can scale these to systems where agents, their harness, and the environment all change together. The challenge is not to make more components adaptive, but to decide which should change, how their updates influence one another, and how to keep the co-evolution pressure productive rather than unstable or dominated by one component. Meta co-evolution begins to address this challenge by letting the system make these evolutionary decisions itself, moving beyond human-predefined paths of improvement.

\paragraph{Safety and governance.}
As co-evolution becomes more autonomous, humans may lose control over the system's behavior. Evolving agents may develop attack strategies, tool-use patterns, communication protocols~\citep{tucker2022trading,motwani2024secret}, or organizational behaviors that exceed human understanding and are hard to monitor. The risk is sharper in meta co-evolution, where systems may also alter which behaviors are rewarded and preserved. Future work therefore needs governance that keeps open-ended evolution auditable and interruptible, including sandboxed deployment, continuous monitoring, rollback to verified states, and human intervention points.

\section{Conclusion}
\label{sec:conclusion}
This survey focuses on co-evolution in agentic systems, where multiple components impose evolutionary pressure on one another. We organize the field through a progressive taxonomy whose adaptive boundary expands from agents alone, to agents together with their environments, and finally to the mechanism governing their joint evolution, reflecting a gradual reduction of human intervention. This perspective shows that future progress lies not in stronger static agents but in agents that continually improve through co-evolution, supported by evaluation and governance that keep the process reliable and controllable.

\section*{Limitations}
\label{sec:limitations}

\paragraph{Meta co-evolution is still at an early stage.}
Only limited work currently meets our definition of Stage~3. Much of our discussion therefore draws on single-entity meta-evolution as a precursor, which shows that evolution mechanisms can themselves evolve but does not couple this change to a lower-level co-evolving system.

\paragraph{Safety and governance are not operationalized.}
We identify safety, monitoring, and human oversight as first-order concerns for increasingly autonomous co-evolution, and point to failure modes specific to co-evolving systems, such as evaluator exploitation, partner overfitting, and diversity collapse. But we only treat these at the level of desiderata, and do not develop concrete safeguards or protocols for them.

\section*{Ethics Statement}
Our paper presents a comprehensive survey of co-evolution in agentic systems, with a specific focus on how agents and their environments adaptively reshape one another, moving from human-designed toward self-directed evolution. All research works reviewed in this survey are properly cited. To the best of our knowledge, the referenced materials are publicly accessible or available under licenses permitting their use for research review. We did not conduct additional dataset curation or human annotation work. Consequently, we believe that this paper does not raise any ethical concerns.

\bibliography{custom}

\newpage
\appendix

\begin{center}
    {\Large\textbf{Appendices}}
\end{center}

\section{Distinguishing Co-Evolution from Adjacent Concepts}
\label{app:adjacent-concepts}

\begin{table*}[t]
\centering
{
\small
\setlength{\tabcolsep}{4pt}
\renewcommand{\arraystretch}{1.13}
  \vspace{-0.15in}
\begin{tabularx}{\textwidth}{@{}p{0.8cm}p{4cm}p{4.4cm}X@{}}
\toprule
\textbf{Survey} & \textbf{Main scope} & \textbf{Original taxonomy} & \textbf{Coverage of co-evolution} \\
\midrule
\citet{meng2026agentharness}
& Runtime infrastructure that governs LLM-agent execution
& Where the harness sits in the agent stack and whether it is general or domain-specific
& \textbf{No dedicated coverage.} The survey is organized around harness components and contains no co-evolution category or section. \\
\addlinespace
\citet{guo2024llmmultiagents}
& Design, communication, and collaboration in LLM-based multi-agent systems
& Agents--Environment Interface; Agents Profiling; Agents Communication; Agents Capability Acquisition
& \textbf{No dedicated coverage.} Section~3.4 lists self-evolution as one strategy for updating agent capabilities, but the survey has no section on co-evolution. \\
\addlinespace
\citet{huang2025envscaling}
& How environments generate tasks, support agent execution, and provide training feedback
& Task Generation; Task Execution; Feedback
& \textbf{Mainly discussed as a future direction.} Section~6 includes ``Co-Evolution via Embedded External Tools,'' but co-evolution is not part of the main taxonomy. \\
\addlinespace
\citet{fang2025selfevolvingaiagents}
& Techniques for self-evolving agentic systems and their domain-specific applications
& System Inputs; Agent System; Environment; Optimisers
& \textbf{Mentioned as an open challenge.} Section~8.1.3 identifies the co-evolution of tools alongside agents as underexplored, but it is not part of the main taxonomy. \\
\addlinespace
\citet{ren2026selfimprovements}
& Self-improvement of the foundation model and its operational scaffold
& Foundation Model Improvement; Scaffolding Improvement
& \textbf{Mainly discussed as a future direction.} Section~9.2 lists Multi-Agent Cooperative Co-Evolution as one research direction, but it is not part of the main taxonomy. \\
\addlinespace
\citet{xie2026surveyharness}
& Architecture and evolution of the infrastructure surrounding an agent model
& Execution \& Orchestration; Context \& Trajectory Management; Interaction Surface \& Execution Environment; Constraints \& Guardrails
& \textbf{Discussed as a cross-cutting design pattern.} It covers model--harness co-evolution, but under our definition, changes to both still constitute the evolution of a single agent. \\
\addlinespace
\citet{chen2026recursive}
& Recursive self-improvement from bounded refinement to autonomous AI research
& Deployment-Time Self-Evolution; Training-Time Self-Iteration; Self-Evaluation; Auto Research, crossed with Human-in-the-loop, Human-on-the-loop, and Closed loop
& \textbf{Covered through scattered examples.} Co-evolution appears in self-play and evaluator evolution, but it is not part of the main taxonomy. \\
\addlinespace
\citet{gao2025selfevolvingagents}
& Self-evolving agents, including single-agent component updates and multi-agent methods
& What, When, How, and Where to Evolve
& \textbf{Covered across several subsections.} Sections~3.4.2, 5.3.2, and 6.1 cover multi-agent and model--agent evolution without a unified co-evolution taxonomy. \\
\addlinespace
\citet{xiang2026systematicselfevolving}
& The broader field of self-evolving agents, including model-centric and environment-centric single-entity evolution
& Model-Centric Self-Evolution; Environment-Centric Self-Evolution; Model--Environment Co-Evolution
& \textbf{Included as one top-level category.} Model--Environment Co-Evolution is one of three branches, but the survey does not cover Agent--Agent or Meta Co-Evolution. \\
\addlinespace
\textbf{Ours}
& Mutual adaptation among components in agentic systems
& Agent--Agent Co-Evolution; Agent--Environment Co-Evolution; Meta Co-Evolution
& \textbf{Co-evolution is the central subject and organizing principle.} We cover adaptive agents, environments, and evolution mechanisms. \\
\bottomrule
\end{tabularx}
}
  \vspace{-0.05in}
\caption{Comparison with surveys in adjacent areas. We retain each survey's original taxonomy and report where co-evolution appears in its organization.}
  \vspace{-0.1in}
\label{tab:related-surveys}
\end{table*}

\paragraph{Multi-agent interaction.}
Multiple agents can communicate, collaborate, or compete without changing themselves~\citep{wu2023autogen,guo2024llmmultiagents}. Interaction becomes co-evolution only when it produces persistent updates to multiple agents and their changes continue to shape one another.

\paragraph{Agent loops and loop engineering.}
An agent loop organizes the repeated steps used to complete a task~\citep{xie2026surveyharness}. Loop engineering designs goals, feedback, verification, and stopping rules of this process. Both concern how tasks are executed rather than whether the agent changes across tasks.

\paragraph{Harness engineering.}
Harness engineering designs the persistent runtime components of an agent, including its prompts, memory, tools, skills, and workflows~\citep{xie2026surveyharness}. Designing a harness alone is not evolution. Evolution occurs only if the model, the harness, or both are updated and retained across runs~\citep{SkillRevise}. Even when both change, they still constitute one evolving agent in our definition.

\paragraph{Evolutionary optimization.}
Evolutionary optimization searches over candidate prompts, programs, or agent designs through generation, evaluation, and selection~\citep{chauhan2025evolutionaryllm}. Its candidates only need to be evaluated and selected, while co-evolution requires persistent adaptive units that change in response to one another.

\paragraph{Continual learning.}
Continual learning repeatedly updates the same learner and preserves its changes as new data or tasks arrive~\citep{parisi2019continual}. Unlike evolutionary optimization, it accumulates changes within one continuing learner rather than selecting among separate candidates. Its learning procedure can remain fixed, and no other component needs to adapt.

\paragraph{Self-play.}
Self-play is a training scheme in which a learner interacts with copies or earlier versions of itself~\citep{silver2017alphagozero,zhang2024selfplay}. It can support self-evolution even when only the current learner changes. In contrast, co-evolution requires multiple adaptive units that retain their changes and continue shaping each other.

\paragraph{Self-evolution.}
Self-evolution allows an agent to improve while its tasks, feedback rules, counterparts, or other learning conditions remain fixed~\citep{gao2025selfevolvingagents}. Co-evolution requires another adaptive unit whose changes reshape the agent's later evolution while the agent also reshapes that unit. This additional unit may be created by the agent itself, but it must evolve as a distinct part during the process.

\paragraph{Meta-evolution.}
Meta-evolution changes the mechanism that governs future evolution, including what, when, how, and where to evolve, and how changes are evaluated~\citep{promptbreeder2023,chen2026recursive}. The meta-level controller may remain fixed as long as it modifies this mechanism. Meta Co-Evolution further requires the modification to be driven by and then reshape a lower-level co-evolving system.

\paragraph{Iterative and recursive self-improvement.}
Iterative self-improvement repeatedly applies a fixed improvement procedure to make a system better. Recursive self-improvement also improves this procedure, so the updated system can make more effective self-improvements in later rounds~\citep{schmidhuber2007godelmachine,chen2026recursive}. Both may occur within a single system and therefore do not necessarily constitute co-evolution.

\paragraph{Open-endedness.}
Open-endedness means that a system continues producing novel and learnable outcomes instead of settling at a fixed endpoint~\citep{stanley2017open,hughes2024open}. It describes a possible long-term outcome rather than a particular form of evolution, and co-evolution does not automatically guarantee it.

\section{Comparison with Related Surveys}
\label{app:survey-comparison}

Table~\ref{tab:related-surveys} compares our survey with adjacent surveys using their original taxonomies. Existing surveys primarily organize the literature around multi-agent systems, agent harnesses, self-evolution, recursive improvement, or environment scaling. Co-evolution consequently appears as scattered examples, a future direction, or one branch of a broader taxonomy. In contrast, we use co-evolution as the central inclusion criterion and distinguish where it occurs across agents, environments, and evolution mechanisms.

\section{Construction of the Cross-Paper Evidence Figure}
\label{app:cross-paper-evidence}

Figure~\ref{fig:cross-paper-evidence} summarizes results reported by prior work rather than results from new experiments. Because the included papers use different models, benchmarks, and evaluation metrics, we do not directly pool their raw scores. Instead, all comparisons and aggregations are first performed within each paper.

\paragraph{Matched comparisons in Panels A and B.}
We include a paper only when it reports a static or frozen counterpart and an evolving counterpart under the same backbone and evaluation setting. When multiple baselines are available, we select the controlled variant that differs from the full method primarily in whether the relevant counterpart is allowed to evolve, rather than selecting an arbitrary or strongest baseline. Evaluation settings without results for both conditions are excluded.

For Panel A, we compute the mean score of the matched settings separately for the static and evolving conditions. Metrics are direction-aligned before averaging, such that a higher value consistently indicates better performance. The two endpoints should therefore be interpreted only as a within-paper comparison; their absolute positions are not directly comparable across papers.

Panel B uses the same matched results. Its horizontal coordinate is the number of matched evaluation settings, and its vertical coordinate is the mean paired improvement produced by enabling counterpart evolution. Bubble area represents the proportion of matched settings with a positive improvement, while color indicates the stage in our taxonomy. No unmatched or aggregate-only result is treated as an individual evaluation setting.

\paragraph{Normalized trajectories in Panel C.}
Panel C is constructed from performance trajectories reported across successive evolution rounds, epochs, or training checkpoints. Values are taken from explicitly reported labels when available and otherwise approximately digitized from the published plots. For each trajectory, evolution progress is normalized to the interval $[0,1]$:
\[
\hat{x}_t = \frac{x_t-x_0}{x_T-x_0}.
\]
Performance is expressed as the fraction of the maximum observed improvement:
\[
\hat{y}_t =
\frac{y_t-y_0}
{\max_{\tau} y_{\tau}-y_0}.
\]
Lower-is-better metrics are direction-aligned before normalization. Trajectories without an observed positive improvement are not included.

Each normalized trajectory is linearly interpolated onto a common progress grid. When a paper reports multiple models or evaluation metrics, these trajectories are first averaged within that paper, so that every paper contributes one equally weighted curve. The solid line in Panel C is the mean of the paper-level trajectories, and the shaded region denotes one sample standard deviation across papers. This band reflects cross-paper heterogeneity rather than experimental uncertainty or a confidence interval.

\end{document}